\documentclass[runningheads]{llncs}
\usepackage{graphicx}
\usepackage{float}
\usepackage{subcaption}
\usepackage{tikz}
\usetikzlibrary{calc}
\usepackage{siunitx}
\usepackage{skmath}
\usepackage{amsmath}
\usepackage{amssymb}
\usepackage{nicefrac}
\usepackage{cleveref}
\makeatletter
\AtBeginDocument{%
	\if@cref@abbrev\crefname{figure}{Fig.}{Fig.}%
	\else\Crefname{figure}{Figure}{Figures}\fi%
}%
\AtBeginDocument{%
	\if@cref@abbrev\crefname{equation}{Eq.}{Eq.}%
	\else\Crefname{equation}{Equation}{Equations}\fi%
}%

\begin{document}
\title{An approach for combining transparency and motion assistance of a lower body exoskeleton}
\titlerunning{Combining exoskeleton transparency and motion assistance}
%
\author{Jakob Ziegler\inst{1}\orcidID{0000-0002-3154-6650} \and
	Bernhard Rameder\inst{1}\orcidID{0000-0002-6792-6129} \and
	Hubert Gattringer\inst{1}\orcidID{0000-0002-8846-9051} \and
	Andreas Müller\inst{1}\orcidID{0000-0001-5033-340X}}
\authorrunning{J. Ziegler et al.}
%
\institute{Institute of Robotics, Johannes Kepler University Linz\\Altenberger Str. 69, 4040 Linz, Austria
	\email{\{jakob.ziegler,bernhard.rameder,hubert.gattringer,a.mueller\}@jku.at}\\
	\url{https://www.jku.at/en/institute-of-robotics}\\
}
\maketitle              
\begin{abstract}
	In this paper, an approach for gait assistance with a lower body exoskeleton is described. Two concepts, transparency and motion assistance, are combined. The transparent mode, where the system is following the user's free motion with a minimum of perceived interaction forces, is realized by exploiting the gear backlash of the actuation units. During walking a superimposed assistance mode applies an additional torque guiding the legs to their estimated future position. The concept of adaptive oscillators is utilized to learn the quasi-periodic signals typical for locomotion. First experiments showed promising results.
	
\end{abstract}
\section{Introduction}
In order to cope with the increasing demand for rehabilitation and movement assistance due to the global demographic change, interest in robotic solutions as supplement to classical methods and therapies is growing tremendously. Naturally, also wearable robotic devices, like exoskeletons, are getting more and more focused on. The most commonly known type of exoskeletons are probably systems assisting the human walking motion that are e.g. applied in the field of robotic gait rehabilitation. Due to the close interaction between the human body and the robotic system, an immense variety of research topics is related to wearable robotics in general and exoskeletons in particular, which renders a detailed and exhaustive overview of the international status of research difficult. However, robustly recognizing the user's intended motion and adapting to it in an intelligent way, as well as the accurate timing of provided assistance can be defined as crucial factors in assistive wearable robotics \cite{Koller2015}.\newline
Considering human gait, the cyclic nature of locomotion patterns is used to learn the current movement and to anticipate the appropriate assistance. To this end, the methodology of adaptive oscillators have already been successfully applied in the past \cite{Aguirre2015,Giovacchini2015,Seo2016}. These publications all focused on assisting the flexion and extension of the hip joints during walking. In \cite{Aguirre2015} adaptive oscillators are used to synchronize the assistive torque to the net muscle torque, which is assumed to be quasi-periodic and estimated with the measured hip angle and average values of the dynamic parameters of leg and hip joint of adult persons. The authors of \cite{Seo2016} used adaptive oscillators paired with non-sinusoidal functions representing nominal joint angle trajectories based on experimental data to synchronize to the current gait phase of the user. The assistance torque is then calculated based on predefined torque patterns as a function of the gait phase.
In \cite{Giovacchini2015}, a modified version of a Hopf oscillator is utilized as adaptive oscillator, which is then coupled with a kernel-based non-linear filter to track and learn the periodic features of the hip angle trajectory. Assistance is realized by calculating the difference between the current position and the estimated future position of the joint based on the learned signal and a phase shift.

In line with above mentioned publications, this paper deals with a methodology towards the assistance of gait with a lower extremity exoskeleton. Adaptive oscillators are used to learn the current quasi-periodic movement, represented by the measured hip angle trajectories, and to anticipate the appropriate assistance. To support this well-known methodology and to increase the overall robustness, artificially generated gait patterns are used to tune the learning gains and to limit the parameters of the estimated signals. Foundation of the overall functionality of the robotic device is the implementation of transparency \cite{Just2018}, where the system follows the user's free motion without hindering it significantly. As a novel approach, this is established in a simple but effective way by exploiting the gear backlash of the actuation units.

\section{Setup}
\begin{figure}[htb]
	\centering
	\begin{subfigure}{0.3\textwidth}
		\centering
		\includegraphics[height=5.3cm]{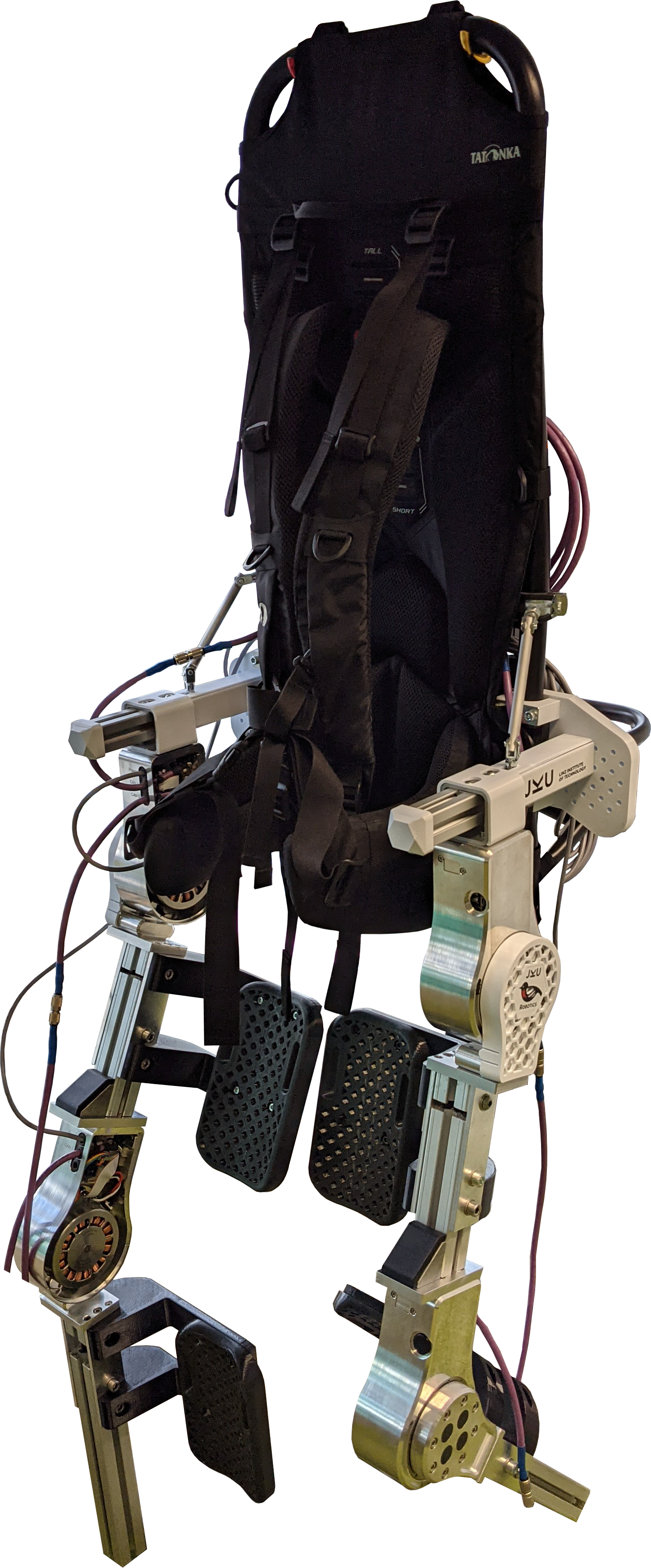}
	\subcaption{hip and knee joints}
	\label{fig:exo_4dof}
	\end{subfigure}
	\hfill
	\begin{subfigure}{0.3\textwidth}
	\centering
	\includegraphics[height=5.3cm]{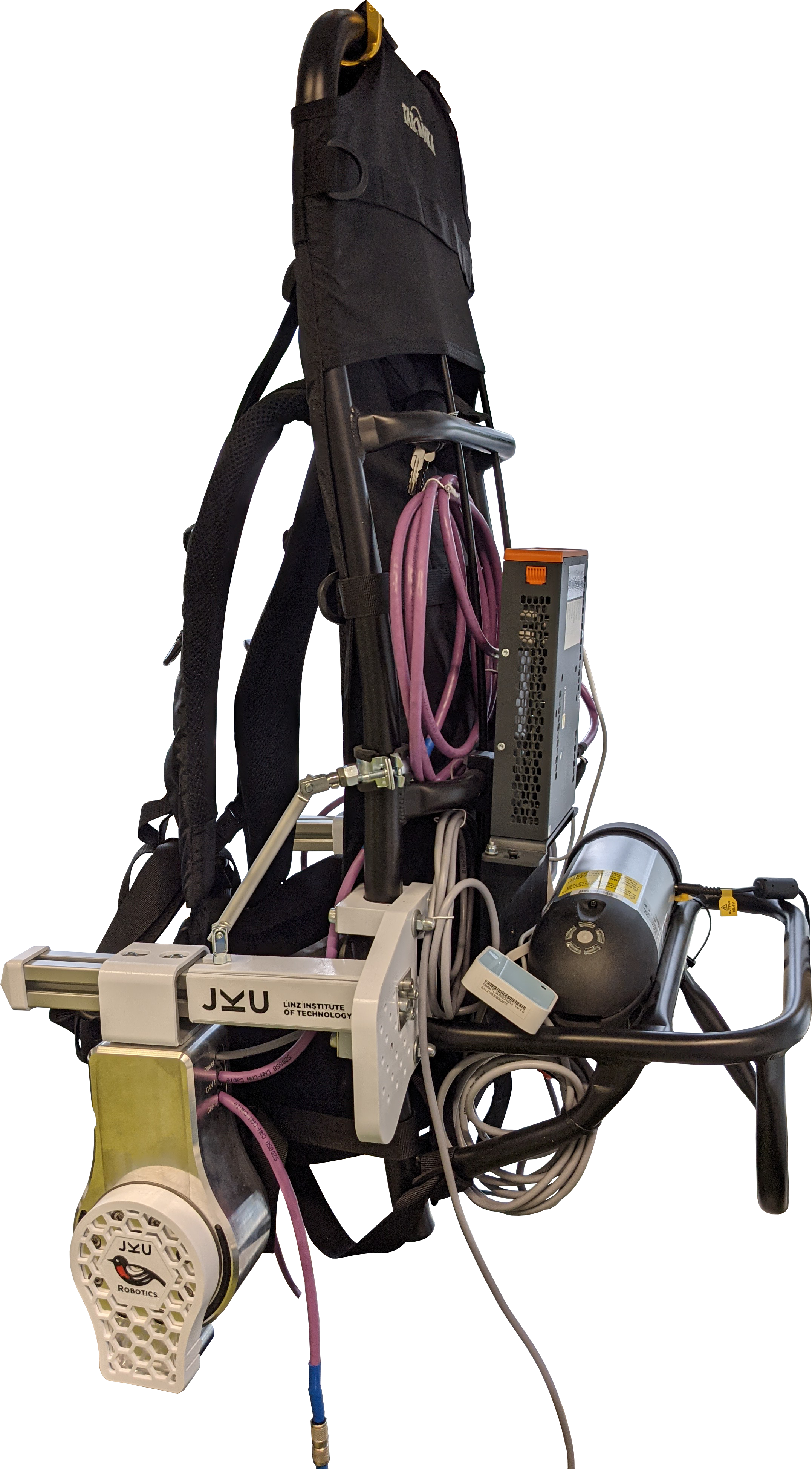}
	\subcaption{hip joints}
	\label{fig:exo_2dof}
	\end{subfigure}
	\hfill		
	\begin{subfigure}{0.37\textwidth}
		\centering
		\includegraphics[height=5.3cm]{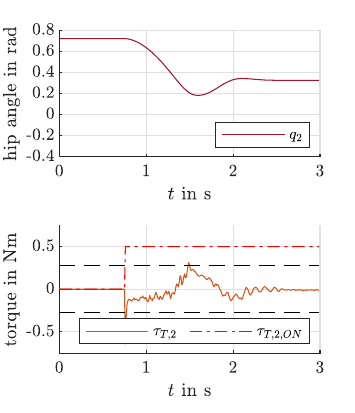}
		\subcaption{transparency mode check}
		\label{fig:exo_TM_Test}
	\end{subfigure}
	\caption{Current setup of the exoskeleton and show-case of transparency mode}
	\label{fig:exo}
\end{figure}

Nowadays several exoskeletons are commercially available. However, a commercial system is typically not suitable for research purposes, as the possibilities to investigate and adapt the implemented methods and algorithms are usually extremely limited. On this account, an exoskeleton prototype is currently being developed at the Institute of Robotics of the Johannes Kepler University Linz. It serves as a testing platform, where novel control strategies, motion assistance approaches and the like can be implemented and tested in a low-level manner with full access to the functionality of the system.\newline
In the current status four actuation units are in operation, two positioned at the hip joints and two at the knee joints of the user. One actuation unit (\textit{Exoskeleton Drive GEN.1}, Maxon, Switzerland) consists of a brushless DC motor with an internal incremental encoder, a control unit, a planetary gear head (gear ratio 448/3) and an absolute encoder at the gear output. 
\newline
The modular conception of the system allows for an easy change between two (hip joints) or four (hip and knee joints) actively driven degrees of freedom. \Cref{fig:exo_4dof} and \ref{fig:exo_2dof} shows the current status of the exoskeleton prototype. For the measurements and experiments described in this paper the version with detached knee actuators was used.

\section{Transparency}
To ensure overall usability of a wearable robotic system it is highly desirable that even when no active assistance is required, the device should not hinder the user's free motion in any way. In consequence, unwanted and disturbing effects, for instance due to friction, inertia or gravity, have to be compensated in order to make the powered but otherwise not assisting device \textit{transparent} to the user. As no force or torque sensors or the like are included in the current setup, this transparency has to be realized with the existing components. In a simple but surprisingly effective approach the gear backlash of the actuators is exploited. Inside this gear backlash, which has an angular range of about \SI{\pm 0.5}{\degree}, the gear output side of the actuators can be moved without the motion being transferred to the electric motor on the gear input side.
Both, the motor position $q_{M,j}$ and the actuator position $q_j$ on the gear output side of actuator $j$ are measured and hence the difference of these positions due to the gear backlash can be calculated. When the gear output is moving without the motion being transferred to the electric motor, this position difference can be used to control the system in a way so that it becomes transparent to the user. The according torque $\tau_{T,j}$ for the $j^{th}$ actuator is calculated as

\begin{equation}
	\tau_{T,j} = K_T(q_j - q_{M,j}),
\end{equation}
where $K_T$ is a tune-able virtual stiffness. Applying this torque establishes the desired behavior by significantly decreasing the otherwise quite noticeable and interfering effects, mainly caused by inertia. In addition to the restriction of $\tau_{T,j}$, due to the limited backlash, the resulting torque of the system was kept within bounds by choosing the virtual stiffnesses in an appropriate way.\\
The functionality of the transparent mode was demonstrated by applying the torque $\tau_{T,j}$ to the hip joint, which was initially deflected by a certain angle, causing it to behave like a pendulum. The joint quickly converged to its equilibrium (\Cref{fig:exo_TM_Test}), which is, due to an eccentrical center of mass, not the vertical. The strong damping of the oscillation is caused by the inertia and friction of the planetary gear.

\section{Motion assistance}
Two critical factors for the support of movement with an actuated prosthetic or orthotic device are the robust and correct recognition of the user's motion intention, as well as the accurate timing of the provided assistance \cite{Koller2015}. Thus, when the device is designed to assist a specific movement task, it has to adapt to the characteristics of the intended motion. Concerning human gait, the motion patterns are characterized by their (quasi-)periodic behavior.

\subsection{Learning of periodic signals}
One method to adapt to a signal with cyclic nature and to extract its relevant features is the use of so-called adaptive oscillators, the mathematical description of a non-linear dynamical system capable of learning periodic signals \cite{Righetti2006,Buchli2008,Ronsse2011}. If the input signal has several frequency components, such an oscillator will generally adapt to the component closest to its initial intrinsic frequency. In \cite{Righetti2006b} it is proposed to use a set of adaptive oscillators, coupled in parallel, to achieve sort of a dynamic Fourier series decomposition of any periodic input signal. Each of the individual oscillators will adapt to the frequency component closest to its intrinsic frequency. If thereby the individual intrinsic frequencies $\omega_i = i\omega$ of the $i=[1, \ldots, N]$ oscillators are defined as multiples of a fundamental frequency $\omega$. The update rules are given by
\begin{align}
	\dot{\phi}_i &= i\omega + \kappa_{\phi}p\cos{\phi_i} \\
	\dot{\omega} &= \kappa_{\omega}p\cos{\phi_1} \\
	\dot{\alpha}_0 &=\kappa_{\alpha}p \\
	\dot{\alpha}_i &= \kappa_{\alpha}p\sin{\phi_i},
\end{align}
where $\phi_i$ is the phase and $\alpha_i$ is the amplitude of the $i^{th}$ oscillator and $\kappa_{\phi}$, $\kappa_{\omega}$ and $\kappa_{\alpha}$ are learning gains. The periodic perturbation $p=q_j-\hat{q}_j$ is defined as the difference between the measured position $q_j$ of the $j^{th}$ actuator and its estimation \cite{Ronsse2011}
\begin{equation}
	\hat{q}_j=\alpha_0+\sum_{i=1}^{N}\alpha_i\sin{\phi_i}.
\end{equation}
The choice of the learning gains is crucial for the overall behavior of this algorithm. A recommendation in this regard is proposed in \cite{Ronsse2013}, based on which the parameters were initially set. To simulate above methodology and to fine tune the parameters, artificially generated gait patterns were used. To this end we used the approach described in a previous publication \cite{Ziegler2022} to synthesize hip angle trajectories according to a simulated person walking at variable gait speed.

\subsection{Assistive torque}
Adaptive oscillators can be used to synchronize to a quasi-periodic input signal, which could be the angular position of joints or even electromyographic measurements of the leg muscles \cite{Aguirre2015,Giovacchini2015,Seo2016}.
In the present case the joint angles of the hip actuators are used. Once the frequency, phase and amplitude components of the input signal and thus its periodic behavior is learned, it can be reproduced based on the current parameters. An estimate of the current joint angle $\hat{q}_j(t)$ and its future value $\hat{q}_j(t+\Delta t)$ according to a specified time horizon $\Delta t$ are therefore provided. Based thereon, an assistive torque $\tau_{A,j}$ can be calculated as
\begin{equation}
	\tau_{A,j} = K_A(\hat{q}_j(t+\Delta t) - \hat{q}_j(t)),
	\label{eq:tau_A}
\end{equation}
where $K_A$ is a tune-able virtual stiffness \cite{Giovacchini2015}. Thus, the limbs of the user are smoothly assisted towards their estimated future position, while the system adapts to changes in the gait pattern in terms of frequency and shape, e.g. due to varying gait speed.\newline
Following this strategy an additional torque of course can only be considered as \textit{assistive} when the reproduced periodic signal resembles the measured signal and the discrepancy between estimated and actual movement is small. To ensure usability an additional weight
\begin{equation}
	w_{A,j} = \frac{1}{2}\left( 1-\tanh{\frac{p-p_{max}}{\epsilon}} \right)
\end{equation}
is defined, where $p(t)$ again is the signal difference between input and estimation, $p_{max}$ adjusts the allowed maximum of this difference and $\epsilon$ tunes the steepness of the $\tanh{}$ function. The final torque
\begin{equation}
	\tau_j = \tau_{T,j} + w_{A,j}\tau_{A,j}
\end{equation}
applied onto the $j^{th}$ actuator is then the combination of the torque components according to transparency and motion assistance.

\section{Experimental Results}
\begin{figure}[htb]
	\centering
	\includegraphics[width=\linewidth]{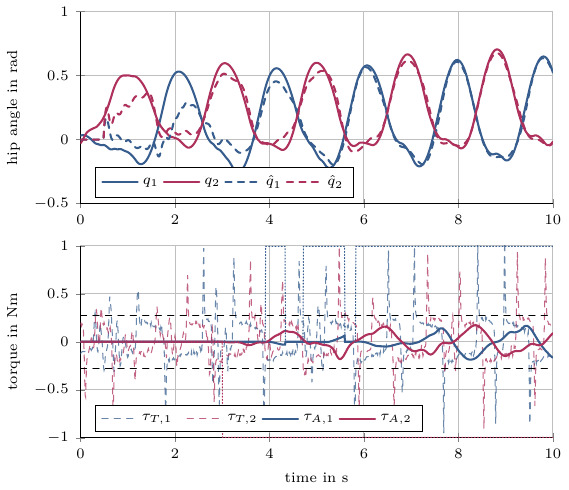}
	\caption{Adaption process with measured (solid) and estimated (dashed) signals on the top and torques according to calculated assistance (solid) and transparency (dashed) on the bottom (activated assistance dotted)}
	\label{fig:fig_TM_AM}
\end{figure}

\begin{figure}
	\centering
	\includegraphics[width=\linewidth]{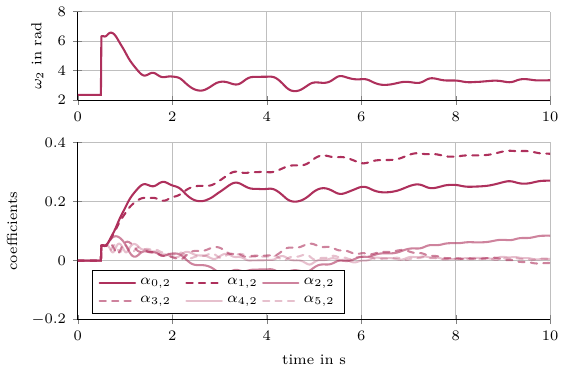}
	\caption{Adaption of the angular frequency (top) and the coefficients (bottom)} 
	\label{fig:fig_omega_alpha}
\end{figure}

First experiments with the described exoskeleton prototype were successfully conducted, where the assistive torque was superimposed on the torque according to the presented transparent mode. These initial tests were performed by persons of our institute walking on level ground. \Cref{fig:fig_TM_AM} shows a typical adaption process during gait, starting with the transition from stand to walk. The measured angles of the right ($q_1$) and left ($q_2$) hip joint are shown as solid lines, their estimations $\hat{q}_1$ and $\hat{q}_2$, respectively, as dashed lines. At about \SI{0.5}{\second} the adaption process is (re-)activated and the adaptive oscillator system is reset to defined initial conditions. As can be seen, the estimated signal of the left leg is close enough to the measured joint angle after one gait cycle and the assistive torque (bottom, solid red line) is smoothly switched on. In contrast, the learned signal of the right leg is not accurate enough during the first gait cycle, causing the assistive torque (bottom, solid blue line) to be switched off again, and the motion assistance starts with the third gait cycle. Activated assistance is shown as lines with dots, where $+1$ corresponds to the right hip joint and $-1$ to the left hip joint. Transparency is constantly activated. The according torques $\tau_{T,j}$ are shown as dashed lines. Due to safety reasons, the total desired torque is limited to \SI{+-0.275}{{\newton\meter}} (black dashed lines). In \cref{fig:fig_omega_alpha} the parameters of the oscillator system of the left leg is shown for the same measurement. After the convergence phase during the first few gait cycles the periodic behavior of the motion signals is learned accurate enough for activating the assistance mode, while the system constantly adapts to the current hip angle trajectories.

\section{Conclusion}
This paper presents an approach to assist gait motion using a lower body exoskeleton, which consists of two main parts. On the one hand, a transparent mode is realized, where the device is not applying any assistance or resistance. Unwanted effects due to friction, inertia or gravity are compensated significantly, minimizing the perceived interaction forces between user and exoskeleton. To this end, the gear backlash of the actuators is exploited, representing a low-cost solution as no additional force or torque sensors, or similar, are necessary. On the other hand, an assistance mode is implemented to support the user while walking. The methodology of adaptive oscillators is utilized to learn the quasi-periodic motion of the legs during gait. Artificially generated gait patterns were used to tune the learning gains and to define bounds for the system parameters to increase robustness. Based on the learned motion an assistive torque is calculated and superimposed to the torque needed to achieve transparency.\newline
The introduced methods were implemented and tested on an exoskeleton prototype. Initial experiments with persons starting to walk from stance showed promising results and the functionality of the approach was confirmed. Ascribable to a lack of usable standard tests, the effects of transparency and assistance were assessed just by subjective evaluation, supported by measurements of actuator positions and torques. Extensive experiments not only considering gait, but also stair ascent and descent and other cyclic movement tasks, as well as designing objective performance measures are in the focus of ongoing research.

\subsection*{Acknowledgement}
This work has been supported by the "LCM - K2 Center for Symbiotic Mechatronics" within the framework of the Austrian COMET-K2 program.

%
%
\bibliographystyle{splncs04}
\bibliography{references.bib}

\end{document}